\PassOptionsToPackage{sort}{natbib}

\documentclass[11pt,a4paper]{article}
\usepackage[hyperref]{acl2021}
\usepackage{times}
\usepackage{graphicx}
\usepackage{latexsym}

\usepackage{microtype}
\usepackage{changes}
\usepackage{booktabs}
\usepackage{xspace}
\usepackage[title]{appendix}
\usepackage{soul}
\aclfinalcopy %

\usepackage{misc_symbols}
\usepackage{multirow}

\newcommand{\bert}[0]{{\bf BT}\xspace}
\newcommand{\distilbert}[0]{{\bf d}\bert}
\newcommand{\roberta}[0]{{\bf RB}\xspace}
\newcommand{\distilroberta}[0]{{\bf d}\roberta}

\newcommand{\firstinit}[0]{{\tt Init\#1}\xspace}
\newcommand{\secondinit}[0]{{\tt Init\#2}\xspace}
\newcommand{\randinit}[0]{{\tt Untrained}\xspace}

\newcommand{\testuntrained}[0]{untrained model test\xspace}
\newcommand{\Testuntrained}[0]{Untrained Model Test\xspace}
\newcommand{\testdiffinit}[0]{different initializations test\xspace}
\newcommand{\Testdiffinit}[0]{Different Initializations Test\xspace}

\title{On the Lack of Robust Interpretability of  Neural Text Classifiers}

\author{Muhammad Bilal Zafar \\
  Amazon \\
  \texttt{zafamuh@amazon.com} \\\And
  Michele Donini \\
  Amazon \\
  \texttt{donini@amazon.com} \\\And
  Dylan Slack\thanks{$\,\,\,$Work done during internship at Amazon.} \\
  University of California, Irvine \\
  \texttt{dslack@uci.edu} \\ \AND
  C\'{e}dric Archambeau \\
  Amazon \\
  \texttt{cedrica@amazon.com} \\\And
  Sanjiv Das \\
  Amazon \&  Santa Clara University \\
  \texttt{sanjivda@amazon.com} \\\And
  Krishnaram Kenthapadi\\
  Amazon \\
  \texttt{kenthk@amazon.com} \\}

\date{}

\begin{document}
\maketitle

\begin{abstract}
With the ever-increasing complexity of neural language models, practitioners have turned to methods for understanding the predictions of these models. One of the most well-adopted approaches for model interpretability is \textit{feature-based interpretability}, i.e., ranking the features in terms of their impact on model predictions. Several prior studies have focused on assessing the fidelity of feature-based interpretability methods, i.e., measuring the impact of dropping the top-ranked features on the model output. However, relatively little work has been conducted on quantifying the robustness of interpretations. In this work, we assess the robustness of interpretations of neural text classifiers, specifically, those based on pretrained Transformer encoders, using two randomization tests. The first compares the interpretations of two models that are identical except for their initializations. The second measures whether the interpretations differ between a model with trained parameters and a model with random parameters. Both tests show surprising deviations from expected behavior, raising questions about the extent of insights that practitioners may draw from interpretations.
\end{abstract}

\section{Introduction} \label{sec:intro}

In recent years, large scale language models like BERT and RoBERTa have helped achieve new state-of-the-art performance on a variety of NLP tasks~\cite{devlin_bert_2019,liu_roberta_2019}. While relying on vast amounts of training data and model capacity has helped increase their accuracy, the reasoning of these models is often hard to comprehend. To this end, several techniques have been proposed to interpret the model predictions.

Perhaps the  most 
widely-adopted class of interpretability approaches is that of  \textit{feature-based interpretability} where the goal is to assign an importance score to each of the input features. These scores are also called \textit{feature attributions}. 
Several methods in this class (e.g., SHAP~\cite{lundberg_unified_2017}, Integrated Gradients~\cite{sundararajan_axiomatic_2017})  possess desirable theoretical properties making them attractive candidates for  interpretability. 

Benchmarking analyses often show that these methods possess \textit{high fidelity}, i.e., removing features marked important by the interpretability method from the input indeed leads to significant change in the model output as expected~\cite{atanasova_diagnostic_2020,lundberg_unified_2017}.

However, relatively few investigations have been carried out to understand the \textit{robustness} of feature attributions.
To explore the robustness, we conduct two tests based on randomization:

\xhdr{\Testdiffinit:}
This test operationalizes the \textit{implementation invariance} property of \citet{sundararajan_axiomatic_2017}. Given an input, it compares the feature attributions between two models that are identical in every aspect---that is, trained with same architecture, with same data, and same learning schedule---except for their randomly chosen initial parameters.
If the predictions generated by these two models are also identical, one would also expect the feature attributions to be the same for such \textit{functionally equivalent} models. If the attributions in two cases are not the same, two users examining the \textit{same input} may deem the same features to have \textit{different importance} based on the model that they are consulting.

\xhdr{\Testuntrained:} This test is similar to the test of \citet{adebayo_sanity_2018}. Given an input, it compares the feature attributions generated on a fully trained model with those on a randomly initialized untrained model. The test evaluates whether feature attributions on a fully trained model differ from the feature attributions computed on an untrained model as one would expect.

We conduct the two tests on a variety of text classification datasets. We quantify the feature attribution similarity using \textit{interpretation infidelity}~\cite{arras_explaining_2016} and \textit{Jaccard similarity}~\cite{tanimoto_elementary_1958}. 
The results suggest that:
    (i)
    Interpretability methods fail the \testdiffinit. In other words, two functionally equivalent models lead to different ranking of feature attributions;
    (ii)
    Interpretability methods fail the \testuntrained, i.e., the fidelity of the interpretability method on an untrained model is better than that of random feature attributions.

These findings may have important implications for how the prediction interpretations are shown to the users of the model, and raise interesting questions about reliance on these interpretations. For instance, if two functionally equivalent models generate different interpretations, to what extent can a user act upon them, e.g., investing in a financial product or not.
We discuss these implications and potential reasons for this behavior in \S\ref{sec:conclusion}.

\vspace{2mm}
\xhdr{Related work.}
Model interpretability has different aspects: local~\citep[e.g.][]{lundberg_unified_2017} vs.
global~\citep[e.g.][]{tan_learning_2018},
feature-based~\citep[e.g.][]{lundberg_unified_2017} vs. 
concept-based~\citep[e.g.][]{kim_interpretability_2018} vs.
hidden representation-based~\citep[][]{li_visualizing_2016}.
See~\citet{gilpin_explaining_2018,guidotti_survey_2018} for an overview.
In this paper, we focus on feature-based interpretability, which is a well-studied and commonly used form~\citep{bhatt_explainable_2020,tjoa_survey_2020}.
Specifically, this class consists of a relatively large number of methods, of which some have been shown to satisfy desirable theoretical properties (e.g., SHAP~\citep{lundberg_unified_2017},
Integrated Gradients~\citep{sundararajan_axiomatic_2017},
and DeepLIFT~\cite{shrikumar_learning_2017}).

There are several important aspects of interpretation robustness. Some prior studies have considered interpretability in the context of \textit{adversarial robustness} where the goal often is to actively fool the model to generate misleading feature attributions. 
See for instance~\citet{dombrowski_explanations_2019,ghorbani_interpretation_2019,slack_fooling_2020,anders_fairwashing_2020}.
In this work, rather than focusing on targeted changes in the input or the model, we explore robustness of feature attribution methods to various kinds of randomizations.

Several prior works have focused on quantifying quality of interpretations. See for instance,~\citet{ribeiro_why_2016,alvarez-melis_towards_2018,hooker_benchmark_2019,meng_automatic_2018,tomsett_sanity_2020,adebayo_debugging_2020,yang_benchmarking_2019,chalasani_concise_2020,chen_robust_2019,lakkaraju_robust_2020}.
Closest to ours is the work of \citet{adebayo_sanity_2018}, which is based on checking the saliency maps of randomly initialized image classification models. However, in contrast to
\citeauthor{adebayo_sanity_2018}, we consider text classification. Moreover, while the analysis of \citeauthor{adebayo_sanity_2018} is largely based on visual inspection, we extend it by considering automatically quantifiable measures. We also  extend the analysis to non-gradient based methods (SHAP).

\section{Setup}

We describe the datasets, models, and interpretability methods considered in our analysis.

\xhdr{Datasets.}
We consider four different datasets covering a range of document lengths and number of label classes.
The datasets are:
(i) \textbf{FPB:} The Financial Phrase Bank dataset~\cite{malo_good_2014} where the task is to classify news headlines into one of three sentiment classes, namely, positive, negative, and neutral.
(ii) \textbf{SST2:} The Stanford Sentiment Treebank 2 dataset~\cite{socher_recursive_2013}. The task is to classify single sentences extracted from movie reviews into positive or negative sentiment classes.
(iii) \textbf{IMDB:} The IMDB movie reviews dataset~\cite{maas_learning_2011}. The task is to classify movie reviews into  positive or negative sentiment classes.
(iv) \textbf{Bios:} The Bios dataset of \citet{de-arteaga_bias_2019}. The task is to classify the profession of a person from their biography.
Table~\ref{table:details_data} in Appendix~\ref{app:datasets} shows detailed dataset statistics.

\subsection{Models}

We consider four pretrained Transformer encoders: BERT (\bert),
RoBERTa (\roberta), DistilBERT (\distilbert), and DistilRoBERTa (\distilroberta). The encoder is followed by a pooling layer to combine individual token embeddings, and a classification head. Appendix~\ref{app:models} describes the detailed architecture, training and hyperparameter tuning details. After training and hyperparameter tuning, the best model is selected based on validation accuracy and is referred to as \firstinit.

\xhdr{\Testdiffinit.}
Recall from \S\ref{sec:intro} that this test involves comparing two identical models trained from different initializations. The second model, henceforth referred to as \secondinit, is trained using the same architecture, hyperparameters and training strategy as \firstinit, but starting from a different set of initial parameters.
Since we start from pretrained encoders, the encoder parameters are not intialized. For each layer in the rest of the model, a set of initial parameters different from those in \firstinit is obtained by calling the parameter initialization method of choice for this layer---\textit{He initialization}~\citep{he_delving_2015} in this case---but with a different random seed.

\xhdr{\Testuntrained.}
Recall 
that this test involves comparing the trained model (\firstinit) with a randomly initialized untrained model, henceforth 
called
\randinit.
To obtain \randinit, we start from the \firstinit, and randomly initialize
the fully connected layers attached on top of the Transformer encoders (the encoder weights are not randomized). The initialization strategy is the same as in \Testdiffinit.

\subsection{Interpretability methods}
\label{sec:interp_methods}

We consider a mix of gradient-based and model agnostic methods. Specifically:
Vanilla Saliency (\textbf{VN}) of~\citet{simonyan_deep_2014}, 
SmoothGrad (\textbf{SG}) of~\citet{smilkov_smoothgrad_2017},
Integrated Gradients (\textbf{IG}) of~\citet{sundararajan_axiomatic_2017},
and KernelSHAP (\textbf{SHP}) of~\citet{lundberg_unified_2017}. 
We also include random feature attribution (\textbf{RND}) which corresponds to each feature being assigned an attribution from the uniform distribution, $\mathcal{U}(0,1)$.
Appendix~\ref{app:interp_params} provides details about the parameters chosen for the interpretability methods.

Given an input text document, we tokenize the text using the tokenizer of the corresponding encoder. Finally, for each input feature (that is, token), the feature attribution of the gradient-based methods is a vector of the same length
as the token input embedding. 
For scalarizing these vector scores, we use the L2-norm strategy of~\citet{arras_explaining_2016} and the Input $\odot$ Gradient strategy of \citet{ding_saliency-driven_2019}.

\subsection{Interpretability Metrics}

To compare the feature attributions by various interpretability methods, we use the following metrics.

\vspace{2mm}
\xhdr{(In)Fidelity:}
Given an input text which has been split into $L$ tokens,
$\mathbf{t} = [t_1, \ldots, t_L]$, 
get the vector $\mathbf{\Psi}(\mathbf{t}) = [\psi(t_1), \ldots, \psi(t_L)]$ of feature attributions of the corresponding tokens using the interpretability method to be evaluated. Drop the features from $\mathbf{t}$ in the decreasing order of attribution score until the model prediction changes from the original prediction (with all tokens present). Infidelity is defined as the \% of features that need to be dropped until the prediction changes.
A better interpretability method is expected to need a lower fraction of features to be dropped until the prediction change.
We simulate feature dropping by replacing the corresponding input token with the model's unknown vocabulary token.

The infidelity metric has appeared in many closely related forms in a number of studies evaluating model interpretability~\cite{arras_explaining_2016,atanasova_diagnostic_2020,fong_understanding_2019,lundberg_unified_2017,samek_evaluating_2017,deyoung_eraser_2020}. All of these forms operate by iteratively hiding features in the order of their importance and measuring the change in the model output, e.g., in predicted class probability, or the predicted label itself. We chose number of tokens to prediction change, which is closely aligned with~\cite{arras_explaining_2016}, due to its simplicity as compared to more involved metrics relying on AUC-style measures~\cite{samek_evaluating_2017,atanasova_diagnostic_2020}.

\vspace{2mm}
\xhdr{Jaccard Similarity:}
It is common to show top few most important features to users as model interpretations. See for instance,~\citet{ribeiro_why_2016} and~\citet{schmidt_quantifying_2019}. In order to measure the similarity between feature attributions generated by different methods, we use the Jaccard@K\% metric. 
Given an input $\mathbf{t}$, 
let $\mathbf{s}_{i}$ be the set of top-K\% tokens, when the tokens are ranked based on their importance as specified by an attribution output $\mathbf{\Psi}_i$. Then, given two attribution outputs $\mathbf{\Psi}_i$ and $\mathbf{\Psi}_j$,  Jaccard@K\% measures the similarity between them as: $J(i,j) = \frac{|\mathbf{s}_i \cap \mathbf{s}_j|}{|\mathbf{s}_i \cup \mathbf{s}_j|}$.
If the top-K\% tokens by the two attributions $\mathbf{\Psi}_i$ and $\mathbf{\Psi}_j$ are the same, then $J(i,j) = 1$. In case of no overlap in the top-K\% tokens, $J(i,j)=0$.

Appendix~\ref{app:jac_compute} shows some examples of Jaccard@K\% computation.
\begin{table}[t]
\centering
\begin{tabular}{lrrrr}
\toprule
{} &  \bert &  \roberta &  \distilbert &  \distilroberta \\
\midrule
\textbf{FPB } &   0.83 &      0.85 &         0.82 &            0.84 \\
\textbf{SST2} &   0.87 &      0.91 &         0.88 &            0.90 \\
\textbf{IMDB} &   0.92 &      0.95 &         0.93 &            0.94 \\
\textbf{Bios} &   0.86 &      0.86 &         0.86 &            0.86 \\

\bottomrule
\end{tabular}
\caption{Test accuracy with \firstinit. For any given dataset, all encoders lead to a similar accuracy.}
\label{table:acc_best}
\end{table}

\begin{table*}[hbt!]
\centering
\small
\begin{tabular}{l|cccc|cccc|cccc|cccc}
\toprule
& 
\multicolumn{4}{|c}{\bf FPB}  &
\multicolumn{4}{|c}{\bf SST2} &
\multicolumn{4}{|c}{\bf IMDB} &
\multicolumn{4}{|c}{\bf Bios} \\
{} &  \bert &  \roberta &  \distilbert &  \distilroberta &  \bert &  \roberta &  \distilbert &  \distilroberta &  \bert &  \roberta &  \distilbert &  \distilroberta &  \bert &  \roberta &  \distilbert &  \distilroberta \\
\midrule
\textbf{VN         } &     68 &        64 &           61 &              59 &     50 &        54 &           53 &              54 &     46 &        53 &           52 &              50 &     28 &        22 &           27 &              25 \\
\textbf{SG         } &     68 &        63 &           61 &              58 &     46 &        50 &           53 &              51 &     45 &        52 &           52 &              48 &     27 &        23 &           27 &              23 \\
\textbf{IG} &     68 &        62 &           61 &              59 &     38 &        48 &           50 &              49 &     39 &        47 &           46 &              47 &     24 &        20 &           17 &              22 \\
\textbf{SHP        } &     60 &        54 &           52 &              43 &     22 &        25 &           30 &              27 &     10 &        19 &           11 &              13 &     15 &        15 &           15 &              14 \\
\textbf{RND        } &     72 &        71 &           68 &              70 &     61 &        67 &           66 &              69 &     58 &        58 &           63 &              66 &     51 &        51 &           49 &              56 \\
\bottomrule
\end{tabular}
\caption{Mean infidelity of different interpretability methods for \firstinit (shown as \%). Lower values are better.}
\label{table:fidelity_best}
\end{table*}

\section{Results}
\label{sec:results}

Table~\ref{table:acc_best} shows the test set accuracy of \firstinit model with different encoders on all the datasets.
For all the datasets, different encoders lead to a very similar  accuracy. 
Tables~\ref{table:acc_randinit}  in Appendix~\ref{app:acc} shows the prediction accuracy for \randinit.

\xhdr{Infidelity.} Table~\ref{table:fidelity_best} shows the infidelity of different interpretability methods on the best performing models (\firstinit). The table shows that:
(i) As expected, the infidelity of all interpretability methods is better than RND;
(ii) SHP provides the best performance, followed by IG;
(iii) For a given dataset, even though different encoders have very similar accuracy (Table~\ref{table:acc_best}), the infidelity of the same interpretability method for different encoders can vary widely, e.g., SHP on  IMDB;
(iv) There is no particularly discernable correlation between the models and their infidelity, for instance, with FPB dataset, the distilled Transformers provides same or lower infidelity as compared to the original counterparts (BERT, RoBERTa), whereas the trend is reversed for SST2 data.

Moreover, gradient-based methods in Table~\ref{table:fidelity_best} use the L2-norm reduction (\S\ref{sec:interp_methods}). Table~\ref{table:fidelity_best_ding} in Appendix~\ref{app:saliency_ding} shows that in most cases, the performance is much worse when using the Input $\odot$ Gradient reduction. Hence, for the rest of the analysis, we only use L2-norm reduction.

\xhdr{\Testdiffinit.} 
Comparing \firstinit and \secondinit in Table~\ref{table:pred_overlap} in Appendix~\ref{app:acc}---two otherwise identical models with only difference being the random initial parameters, shows that a vast majority of predictions are common between the two models: meaning that the \textit{two models are almost functionally equivalent}.

\begin{figure}[t]
    \centering
    \includegraphics[width=0.9\columnwidth]{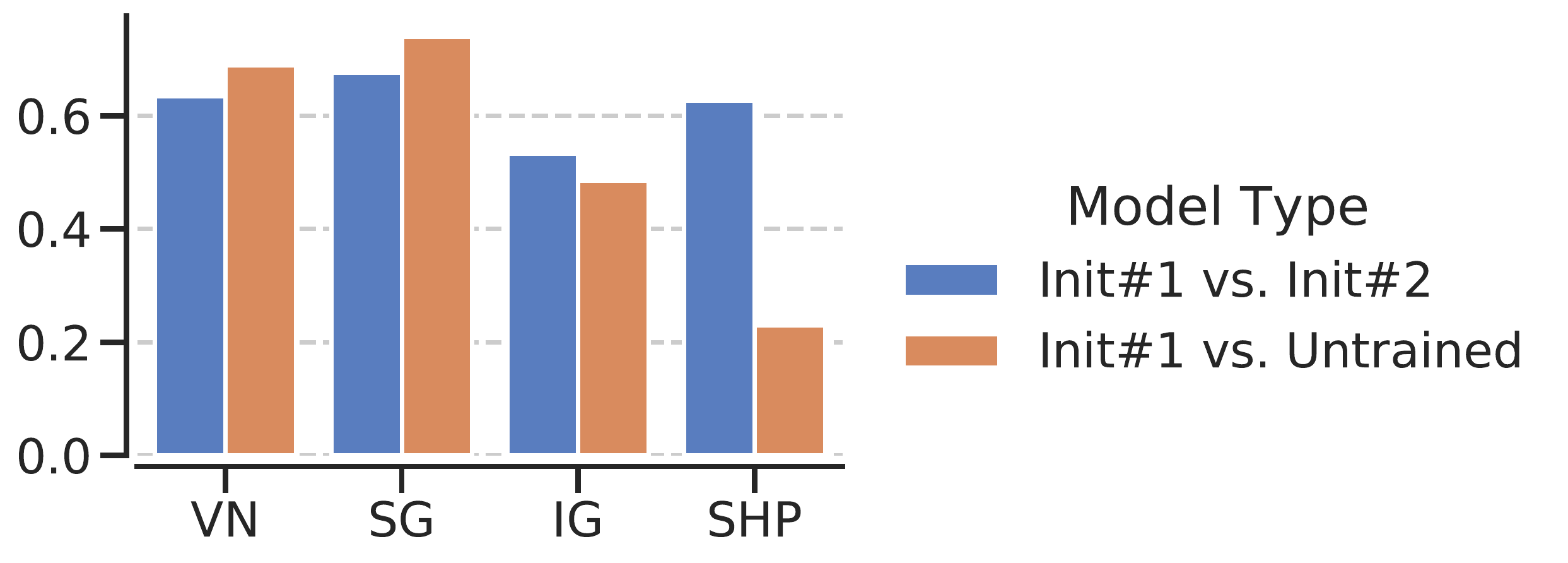}
    \caption{[\bert on SST2 data] Comparing the mean Jaccard@25 between different model types (\firstinit vs. \secondinit and \firstinit vs. \randinit). }
    \label{fig:jacc_comparison_sst2}
\end{figure}

We now compare the similarity in feature attributions of two functionally equivalent models. 
Since the feature attributions are generated w.r.t. the predicted class, our similarity analysis is limited to samples where both models generate the same prediction.
Figure~\ref{fig:jacc_comparison_sst2} shows  Jaccard@25\% when comparing the feature attributions of \firstinit vs. \secondinit, and \firstinit vs. \randinit.
The figure shows that 
(i) for the functionally equivalent models \firstinit vs. \secondinit, Jaccard@25\% is far from the ideal value of $1.0$ --- in fact, for IG, the value drops to almost $0.5$;
(ii) When comparing the top-ranked feature attributions of \firstinit vs. \secondinit, and \firstinit vs. \randinit, the former should show a much bigger overlap than latter,
but this is not the case, except for SHP.

\begin{figure}[t]
    \centering
    \includegraphics[width=0.9\columnwidth]{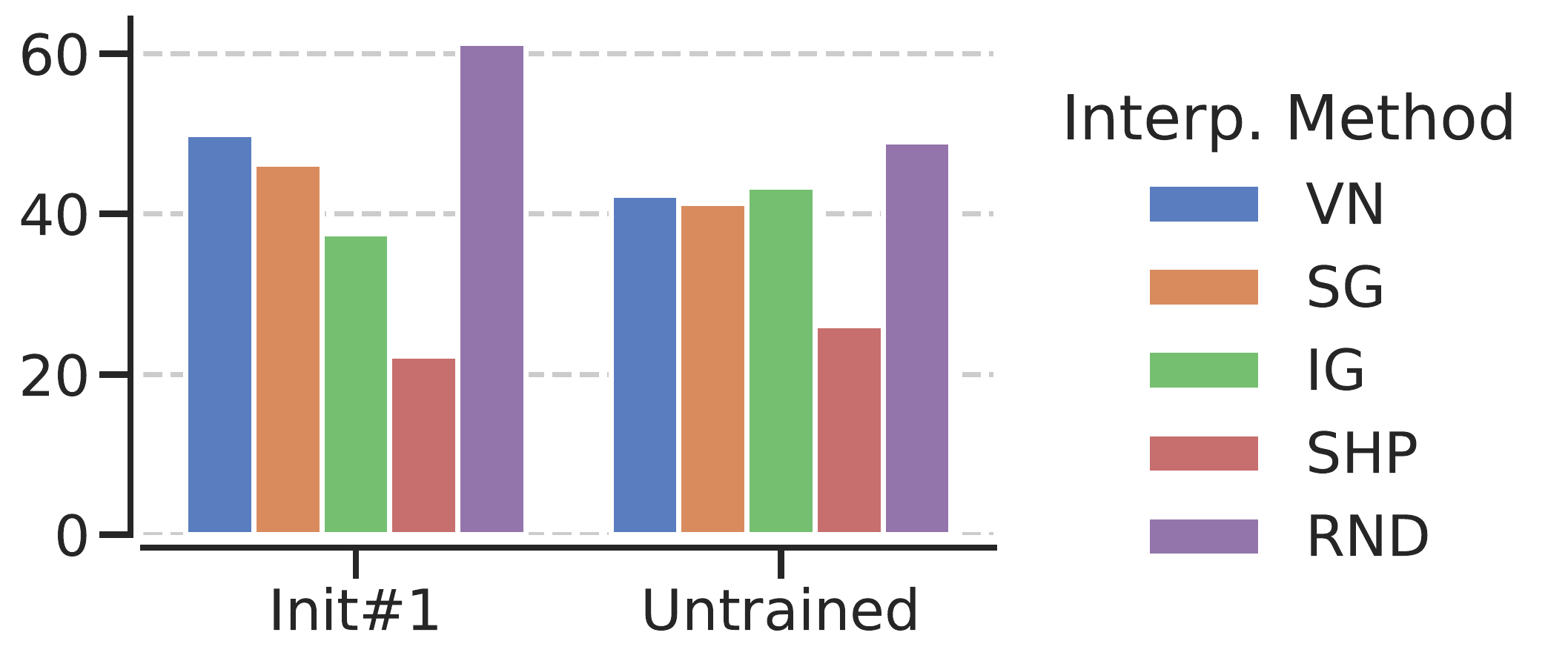}
    \caption{[\bert on SST2 data] Mean infidelity of  interpretability methods on \firstinit and \randinit.
    }
    \label{fig:fidelity_comparison_sst2}
\end{figure}

\begin{table*}[h!t]
\centering
\small
\begin{tabular}{l|cccc|cccc|cccc|cccc}
\toprule
& 
\multicolumn{4}{|c}{\bf FPB}  &
\multicolumn{4}{|c}{\bf SST2} &
\multicolumn{4}{|c}{\bf IMDB} &
\multicolumn{4}{|c}{\bf Bios} \\
{} &  \bert &  \roberta &  \distilbert &  \distilroberta &  \bert &  \roberta &  \distilbert &  \distilroberta &  \bert &  \roberta &  \distilbert &  \distilroberta &  \bert &  \roberta &  \distilbert &  \distilroberta \\
\midrule
\textbf{VN } &                 65 &            52 &                       77 &                  67 &                 63 &            53 &                       73 &                  63 &                 72 &            47 &                       78 &                  72 &                 76 &            70 &                       81 &                  75 \\
\textbf{SG } &                 53 &            66 &                       70 &                  62 &                 68 &            63 &                       70 &                  66 &                 57 &            45 &                       76 &                  75 &                 75 &            70 &                       81 &                  78 \\
\textbf{IG } &                 46 &            35 &                       68 &                  51 &                 53 &            37 &                       71 &                  50 &                 65 &            36 &                       82 &                  68 &                 49 &            51 &                       73 &                  68 \\
\textbf{SHP} &                 57 &            55 &                       64 &                  55 &                 63 &            65 &                       66 &                  62 &                 68 &            50 &                       75 &                  67 &                 72 &            71 &                       75 &                  69 \\
\bottomrule
\end{tabular}
\caption{Jaccard@25\% between the feature attributions for \firstinit vs. \secondinit models (shown as \%).}
\label{table:jaccard_diff_rand_inits}
\end{table*}

\begin{table*}[h!t]
\centering
\small
\begin{tabular}{l|cccc|cccc|cccc|cccc}
\toprule
& 
\multicolumn{4}{|c}{\bf FPB}  &
\multicolumn{4}{|c}{\bf SST2} &
\multicolumn{4}{|c}{\bf IMDB} &
\multicolumn{4}{|c}{\bf Bios} \\
{} &  \bert &  \roberta &  \distilbert &  \distilroberta &  \bert &  \roberta &  \distilbert &  \distilroberta &  \bert &  \roberta &  \distilbert &  \distilroberta &  \bert &  \roberta &  \distilbert &  \distilroberta \\
\midrule
\textbf{VN } &                 65 &            60 &                       75 &                  71 &                 69 &            63 &                       73 &                  69 &                 64 &            79 &                       71 &                  66 &                 71 &            68 &                       70 &                  74 \\
\textbf{SG } &                 47 &            70 &                       67 &                  61 &                 74 &            72 &                       70 &                  75 &                 44 &            53 &                       63 &                  52 &                 66 &            71 &                       60 &                  75 \\
\textbf{IG } &                 41 &            40 &                       55 &                  49 &                 49 &            45 &                       60 &                  49 &                 37 &            75 &                       63 &                  51 &                 40 &            41 &                       55 &                  47 \\
\textbf{SHP} &                 25 &            30 &                       19 &                  29 &                 23 &            22 &                       14 &                  33 &                 15 &            85 &                       23 &                  18 &                 32 &            45 &                       24 &                  27 \\
\bottomrule
\end{tabular}
\caption{Jaccard@25\% between the feature attributions for \firstinit vs. \randinit models (shown as \%).}
\label{table:jaccard_trained_vs_untrained}
\end{table*}

Tables~\ref{table:jaccard_diff_rand_inits} and \ref{table:jaccard_trained_vs_untrained}  show the results for rest of the cases, revealing similar insights.
Specifically, the Jaccard@25\% between \firstinit and \randinit for VN averaged over all 16 cases (four datasets, four models in Table~\ref{table:jaccard_diff_rand_inits}) is 68, whereas the same is 69 when comparing \firstinit and \randinit (Table~\ref{table:jaccard_trained_vs_untrained}). The same comparison yields 67 vs. 64 for SG, 56 vs. 50 for IG and 65 vs. 29 for SHP.
Moving beyond averages, we also counted for each of the cases in Table~\ref{table:jaccard_diff_rand_inits}, the number of times Jaccard@25\% between \firstinit and \secondinit is within 10 units (Jaccard@25\% ranges from 0-100) of the Jaccard@25\% between \firstinit and \randinit for the corresponding pair in Table~\ref{table:jaccard_trained_vs_untrained}. The numbers are 14/16 for VN, 12/16 for SG, 7/16 for IG and 0/16 for SHP. 

We selected K=25 as it corresponds to comparing the 25\% most important features between the two models. Selecting K=10 (comparing 10\% most important features) leads to similar outcomes: 13/16 for VN, 11/16 for SG, 6/16 for IG and 0/16 for SHP.

In other words, \textit{the attribution overlap between two functionally equivalent models can be
similar to
that between a trained vs. an untrained model.}

\vspace{2mm}
\xhdr{\Testuntrained.} Figure~\ref{fig:fidelity_comparison_sst2} shows the infidelity of different methods on SST2 dataset with a \bert model. We note that the performance of RND is better (lower infidelity)
for the untrained model (\randinit) than for the trained model (\firstinit).
Furthermore, even for the untrained model (\randinit), all interpretability methods have a better fidelity than RND. In fact, for SHP, the infidelity is almost half of RND. 
Table~\ref{table:fidelity_rand} in Appendix~\ref{app:additional_results} shows a similar pattern for the rest of the datasets and models.

In short, \textit{%
even for an untrained model, the interpretability methods lead to better-than-random-attribution fidelity.}
The insights highlight the need for baselining the fidelity metric with untrained models before using it as an evaluation measure.

\section{Conclusion \& Future Work}
\label{sec:conclusion}

We carried out two tests to assess robustness of several popular interpretability methods on Transformer-based text classifiers. The results show that both gradient-based 
and model-agnostic methods
can fail the
tests.

These observations raise several \textbf{interesting questions}: if the fidelity of the interpretations is reasonably high on even an untrained model, to what extent does the interpretability method reflect the data-specific vs. data-independent behavior of the model?  If two functionally equivalent models lead to different feature attributions, to what extent can the practitioners rely upon these interpretations to make consequential decisions?

One cause of the non-robust behavior could be the redundancy in text where several input tokens may provide evidence for the same class (e.g., several words in input review praising the movie).
Another reason, related to the first, could be the pathologies of neural models where dropping most of the input features could still lead to highly confident predictions~\citep{feng_pathologies_2018}.%
\footnote{Note, however, that the insights of \citeauthor{feng_pathologies_2018} relate to \textit{dropping least important tokens first}, whereas when computing infidelity, one drops the most important first.}
Dropping individual features can also lead to out-of-distribution samples, further limiting the effectiveness of methods and metrics that rely on simulating feature removal~\cite{kumar_problems_2020,sundararajan_many_2020}.
Systematically analyzing the root causes, and designing interpretability measures that are cognizant of the specific characteristics of text data---preferably with human involvement~\citep{schmidt_quantifying_2019,chang_reading_2009,doshi-velez_towards_2017,poursabzi-sangdeh_manipulating_2021,hase_evaluating_2020,nguyen_comparing_2018}---is a promising research direction. 
Similarly, extending the \Testuntrained to study the effect of randomization of pre-trained embedding models on interpretability is another direction for exploration.

\bibliographystyle{acl_natbib}
\bibliography{NLP_XAI}

\clearpage
\newpage
\appendix

\begin{appendices}

\section{Datasets} \label{app:datasets}
\begin{table}[h]
\begin{tabular}{lrrrr}
\toprule
{} &  $N$ & $K$  &  $D$ &  $\Delta_K$  \\
\midrule
\textbf{FPB } &  4,846   &  3  &  22$\pm$10   & 47 \\
\textbf{SST2} &  9,613   &  2  &  19$\pm$9    & 3  \\       
\textbf{IMDB} &  50,000  &  2  &  227$\pm$168 & 0  \\       
\textbf{Bios} &  397,907 &  28 &  61$\pm$28   & 30 \\      
\bottomrule
\end{tabular}
\caption{Dataset details. The columns are: 
\# of samples ($N$), 
\# of classes ($K$), 
average $\pm$ standard deviation \# of words per document ($D$),
and Class Imbalance ($\Delta_K$). 
Class imbalance is measured as \% prevalence of the most prevalent minus the least prevalent class.
}
\label{table:details_data}
\end{table}

\section{Reproducibility}

\subsection{Architecture, data \& training details}
\label{app:models}

We insert a classification head on top of the pretrained encoder. The end-to-end classifier has the following architecture: 
Encoder $\to$ 
Avg. pooling $\to$  
FC-layer ($512$-units) $\to$ 
RELU $\to$ 
FC-layer ($K$ units), where $K$ is the number of classes.
The maximum sequence length of the encoder is set to $128$ for FPB and SST2 datasets, $512$ for IMDB reviews and $200$ for the Bios data.

Each dataset is split into a $80\%-20\%$ train-test set. $10\%$ of the training set is used as a validation set for hyperparameter optimization. Accuracy and overlap statistics are reported on the test set. 

We used the following hyperparameter ranges: learning rate $\{10^{-2}, 10^{-3}, 10^{-4}, 10^{-5}\}$
and the number of last encoder layers to be fine-tuned $\{0, 2\}$.
Fine tuning last few layers of the encoder, as opposed to  all the layers, has been shown to lead to superior test set performance~\cite{sun_how_2019}.

We use the AdamW optimizer~\cite{loshchilov_decoupled_2019}. 
The maximum number of training epochs is  $25$. We use early stopping with a patience of $5$ epochs: if the validation accuracy does not increase for $5$ consecutive epochs, we stop the training. 
The model training was done using PyTorch~\cite{paszke_pytorch_2019} and HuggingFace Transformers~\cite{wolf_transformers_2020} libraries.

\subsection{Interpretability methods implementation}
\label{app:interp_params}

Owing to the large runtime of methods like SHAP and Integrated Gradients, interpretations are only computed for a randomly chosen $1000$ subsample from the test set. Consequently, metrics like fidelity and Jaccard@K\% are reported only on this subset.

Vanilla Saliency and SmoothGrad are implemented using the PyTorch \texttt{autograd} function. Integrated Gradients and SHAP are implemented using Captum~\cite{kokhlikyan_captum_2020}.
The parameters of these methods are:

\begin{table}[t]
\centering
\begin{tabular}{lrrrr}
\toprule
{} &  \bert &  \roberta &  \distilbert &  \distilroberta \\
\midrule
\textbf{FPB } &    88 &    94 &    90 &    89 \\
\textbf{SST2} &    94 &    96 &    93 &    94 \\
\textbf{IMDB} &    98 &    97 &    97 &    98 \\
\textbf{Bios} &    95 &    95 &    95 &    95 \\
\bottomrule
\end{tabular}
\caption{A vast percentage of predictions are common between the \firstinit and \secondinit, indicating that the models are almost \textit{functionally equivalent}.}
\label{table:pred_overlap}
\end{table}

\begin{table}[t]
\centering
\begin{tabular}{lrrrr}
\toprule
{} &  \bert &  \roberta &  \distilbert &  \distilroberta \\
\midrule
\textbf{FPB } &   0.19 &      0.19 &         0.34 &            0.34 \\
\textbf{SST2} &   0.21 &      0.22 &         0.30 &            0.80 \\
\textbf{IMDB} &   0.27 &      0.51 &         0.20 &            0.46 \\
\textbf{Bios} &   0.03 &      0.02 &         0.01 &            0.01 \\
\bottomrule
\end{tabular}
\caption{Test accuracy with \randinit.}
\label{table:acc_randinit}
\end{table}

\begin{table*}[ht]
\centering
\small
\begin{tabular}{l|cccc|cccc|cccc|cccc}
\toprule
& 
\multicolumn{4}{|c}{\bf FPB}  &
\multicolumn{4}{|c}{\bf SST2} &
\multicolumn{4}{|c}{\bf IMDB} &
\multicolumn{4}{|c}{\bf Bios} \\
{} &  \bert &  \roberta &  \distilbert &  \distilroberta &  \bert &  \roberta &  \distilbert &  \distilroberta &  \bert &  \roberta &  \distilbert &  \distilroberta &  \bert &  \roberta &  \distilbert &  \distilroberta \\
\midrule
\textbf{VN       } &     70 &        70 &           65 &              71 &     62 &        67 &           57 &              70 &     60 &        58 &           51 &              69 &     52 &        53 &           52 &              58 \\
\textbf{SG       } &     70 &        69 &           64 &              69 &     61 &        67 &           54 &              68 &     60 &        58 &           45 &              68 &     51 &        51 &           47 &              58 \\
\textbf{IG} &     75 &        73 &           72 &              71 &     71 &        70 &           65 &              66 &     74 &        64 &           68 &              70 &     59 &        55 &           48 &              58 \\
\bottomrule
\end{tabular}
\caption{Mean infidelity of gradient-based interpretability methods when considering \textit{Input $\odot$ gradient} reduction of~\citet{ding_saliency-driven_2019} for \firstinit (shown as \%). In most cases, the performance is worse than when using the L2-norm reduction (Table~\ref{table:fidelity_best} in Section~\ref{sec:results}). Lower values are better.}
\label{table:fidelity_best_ding}
\end{table*}

\begin{table*}[ht]
\centering
\small
\begin{tabular}{l|cccc|cccc|cccc|cccc}
\toprule
& 
\multicolumn{4}{|c}{\bf FPB}  &
\multicolumn{4}{|c}{\bf SST2} &
\multicolumn{4}{|c}{\bf IMDB} &
\multicolumn{4}{|c}{\bf Bios} \\
{} &  \bert &  \roberta &  \distilbert &  \distilroberta &  \bert &  \roberta &  \distilbert &  \distilroberta &  \bert &  \roberta &  \distilbert &  \distilroberta &  \bert &  \roberta &  \distilbert &  \distilroberta \\
\midrule
\textbf{VN } &     75 &        30 &           33 &              51 &     42 &        43 &           36 &              33 &     35 &        99 &           42 &              43 &     14 &        14 &           11 &              12 \\
\textbf{SG } &     74 &        30 &           33 &              51 &     41 &        40 &           36 &              31 &     36 &        99 &           42 &              43 &     13 &        15 &           11 &              12 \\
\textbf{IG } &     73 &        32 &           31 &              55 &     43 &        42 &           31 &              35 &     49 &       100 &           39 &              42 &     17 &        17 &           11 &              13 \\
\textbf{SHP} &     56 &        19 &           23 &              36 &     26 &        40 &           23 &              27 &      9 &        99 &           35 &              19 &      9 &        11 &            7 &               9 \\
\textbf{RND} &     73 &        36 &           43 &              62 &     49 &        49 &           43 &              51 &     42 &       100 &           49 &              46 &     25 &        32 &           21 &              27 \\
\bottomrule
\end{tabular}
\caption{Mean infidelity of different interpretability methods for \randinit (shown as \%). Lower values are better.}
\label{table:fidelity_rand}
\end{table*}

\xhdr{SmoothGrad.} 
Requires two parameters.
(i) Number of iterations: Following AllenNLP Interpret~\citep{wallace_allennlp_2019}, we use a value of 10.
(ii) Variance of the Gaussian noise $\mathcal{N}(0, \sigma^2)$. The default value of $0.01$ leads to attributions that are almost identical to Vanilla Saliency. So we try different values of $\sigma \in \{0.01, 0.05, 0.1, 0.2\}$ and select the one with the lowest infidelity.

\xhdr{Integrated Gradients.} Requires setting two parameters.
(i) Number of iterations: we use Captum~\cite{kokhlikyan_captum_2020} default of $50$.
(ii) Feature Baseline:  IG requires specifying a baseline~\cite{sundararajan_axiomatic_2017} that has the same dimensionality as the model input, but consists of `non-informative' feature values. We construct the baseline by computing the embedding value of the unknown vocabulary token and repeating it $N$ times where $N$ is the maximum sequence length of the model.

\xhdr{KernelSHAP.} 
Requires two parameters.
(i) Number of feature coalitions: Following the author implementation,%
\footnote{\url{https://github.com/slundberg/shap}}
we use a value of $2L + 2^{11}$, where $L$ is the number of input tokens in the text.
(ii) Dropped token value: SHAP operates by dropping subsets of tokens and estimating model output on these perturbed inputs. We simulate dropping of a token by replacing its embedding value with that of the unknown vocabulary token.

\section{Additional results}
\label{app:additional_results}

\subsection{Accuracy \& prediction commonality}
\label{app:acc}

Table~\ref{table:pred_overlap} shows the fraction of predictions common between \firstinit and \secondinit.

Table~\ref{table:acc_randinit} shows the accuracy of \randinit. As expected, the accuracy of \randinit is much smaller than with trained models.

\subsection{Input $\odot$ Gradient reduction}
\label{app:saliency_ding}
Table~\ref{table:fidelity_best_ding} shows the infidelity of gradient-based interpretability methods when using the Input $\odot$ Gradient dot product reduction of \citet{ding_saliency-driven_2019}. When comparing the results to those with L2 reduction in Table~\ref{table:fidelity_best}, we notice that in \textit{all except two cases} (VN and SG on \distilbert with IMDB data), the performance is worse.

\subsection{Infidelity with \randinit model}

Table~\ref{table:fidelity_rand} shows the infidelity for the \randinit model. Much like Figure~\ref{fig:fidelity_comparison_sst2}, the table shows that in several cases, the performance of the feature attribution methods (most notably SHP) can be much better than random attribution (RND).

The table also shows an exception for \distilbert on IMDB dataset where for all methods, the infidelity is near 100. This behavior is likely an artefact of the particular initial parameters due to which the model always predicts a certain class irrespective of the input.

\section{Examples of top-ranked tokens}
\label{app:jac_compute}

We now show some examples of Jaccard@K\% computation. The examples show the input text, different models, and top-K\% tokens ranked w.r.t. their importance. The attribution method used was VN.

\vspace{2mm}
\xhdr{Example 1: } SST2 data. Comparing \firstinit and \secondinit. Both models predict the sentiment to be positive.

\noindent
\textit{Text.} at heart the movie is a deftly wrought suspense yarn whose richer shadings work as coloring rather than substance

\noindent
\textit{Top-25\% w.r.t. \firstinit.} \{`substance', `rather', `at', `yarn', `coloring', `movie'\}

\noindent
\textit{Top-25\% w.r.t. \secondinit.} \{`heart', `\#\#tly', `suspense', `at', `yarn', `def'\}

\noindent
\textit{Jaccard@25\%.} 20

\vspace{2mm}
\xhdr{Example 2: } SST2 data. Comparing \firstinit and \secondinit. Both models predict the sentiment to be positive.

\noindent
\textit{Text.} an infectious cultural fable with a tasty balance of family drama and frenetic comedy

\noindent
\textit{Top-25\% w.r.t. \firstinit.} \{`fable', `infectious', `cultural', `balance', `an'\}

\noindent
\textit{Top-25\% w.r.t. \secondinit.} \{`cultural', `balance', `infectious', `fable', `an'\}

\noindent
\textit{Jaccard@25\%.} 100

\vspace{2mm}
\xhdr{Example 3: } FPB data. Comparing \firstinit and \randinit. Both models predict the sentiment to be negative.

\noindent
\textit{Text.} nokia shares hit 13.21 euros on friday , down 50 percent from the start of the year in part because of the slow introduction of touch-screen models

\noindent
\textit{Top-25\% w.r.t. \firstinit.} \{ `,', `.', `down', `friday', `shares', `euros', `nokia', `hit' \}

\noindent
\textit{Top-25\% w.r.t. \secondinit.} \{ `,', `.', `down', `euros', `friday', `hit', `shares',  `nokia' \}

\noindent
\textit{Jaccard@25\%.} 100

 In second and third examples, even though the rankings are different, the set of top-25\% tokens is the same leading to a perfect Jaccard@25\%.

\end{appendices}

\end{document}